\documentclass[conference,10pt]{IEEEtran}
\IEEEoverridecommandlockouts

\usepackage{amsmath,amssymb,amsthm}
\usepackage{bm}
\usepackage{mathtools}
\usepackage{dsfont} 

\usepackage{graphicx}
\usepackage{booktabs}
\usepackage{multirow}
\usepackage{makecell}
\usepackage[caption=false,font=footnotesize]{subfig}

\usepackage{algorithm}
\usepackage{algpseudocode}

\usepackage[hyphens]{url}
\usepackage{xcolor}
\usepackage[colorlinks,citecolor=blue,linkcolor=blue,urlcolor=blue]{hyperref}
\usepackage[noabbrev,capitalize]{cleveref}
\usepackage{flushend}

\usepackage[numbers,sort&compress]{natbib}

\newcommand{\vv}[1]{\bm{#1}}
\newcommand{\mm}[1]{\bm{\mathrm{#1}}}
\newcommand{\reals}{\mathbb{R}}
\newcommand{\zz}{\vv{z}}
\newcommand{\xx}{\vv{x}}

\newcommand{\dd}{\vv{d}}
\newcommand{\ww}{\vv{w}}
\newcommand{\bbeta}{\vv{\beta}}
\newcommand{\softmax}{\operatorname{softmax}}
\newcommand{\SAM}{\operatorname{SAM}}
\newcommand{\BCE}{\operatorname{BCE}}
\DeclareMathOperator{\Prob}{\mathbb{P}}
\DeclareMathOperator{\Expect}{\mathbb{E}}
\newcommand{\dicts}{\text{dicts}}
\newcommand{\nnz}{\text{nnz}}
\newcommand{\dimst}{\text{dims}}
\newcommand{\kp}{\text{kp}}
\newcommand{\patches}{\text{patches}}
\newcommand{\dictssel}{\text{dicts,sel}}
\newcommand{\viz}{\text{viz}}

\crefname{section}{Sec.}{Secs.}
\crefname{equation}{Eq.}{Eqs.}
\crefname{figure}{Fig.}{Figs.}
\crefname{table}{Table}{Tables}

\begin{document}
\bstctlcite{IEEEexample:BSTcontrol}

\title{Mechanistic Interpretability for Learning Assurance of a Vision-Based Landing System}

\author{\IEEEauthorblockN{Romeo Valentin\IEEEauthorrefmark{1}\IEEEauthorrefmark{2}, Olivia Beyer Bruvik\IEEEauthorrefmark{1}, Marc R. Schlichting\IEEEauthorrefmark{1}, and Mykel J. Kochenderfer\IEEEauthorrefmark{1}}
\IEEEauthorblockA{\IEEEauthorrefmark{1}Stanford Intelligent Systems Laboratory, Stanford University, Stanford, CA, USA\\
\IEEEauthorrefmark{2}Correspondence to: romeov@stanford.edu\\
}}

\maketitle

\begin{abstract}
EASA's learning-assurance guidance requires data-driven aviation systems
to build and monitor their own \emph{situation representation}, yet for
neural networks the technical means to provide such evidence remain an
open problem. We address this gap for a vision-based aircraft landing
system: we propose that a minimally assurable model must at least be
shown to separate \emph{content} from \emph{style} in its own situation
representation. Showing that the model's predictions then rely largely
on the contentful representation components leads to a concrete
assurance path. To demonstrate this assurance path on a concrete model
we train a vision transformer model for runway keypoint regression on
the LARDv2 dataset. The model, which acts as the subject for our
assurance demonstration, produces per-patch embeddings that we
decompose into interpretable atoms via K-SVD sparse dictionary
learning. A qualitative visualization confirms that contentful atoms
track task-relevant runway structure and stylistic atoms track
domain-specific appearance, and the regression head is shown to place
almost all of its linear weight on contentful atoms. Having validated
the model's situation representation at test time, we further build on
the content/style separation and define \emph{out-of-model-scope}
(OOMS) detection, a novel runtime assurance approach directly
monitoring the model's situation representation. OOMS monitoring is
complementary to operational design domain and output-space
out-of-distribution monitoring and addresses concrete requirements of
the recent EASA guidance. We operationalize OOMS via a constrained
$L_1$ logistic-regression classifier over binary atom activations
derived from the activation of interpretable atoms found earlier. The
model achieves AUROC $0.96$ on an artificial dataset derived from
LARDv2 by considering only $34$ of $512$ atoms, with $78\%$ of the
selected atoms being contentful. By directly analyzing a model's
situation representation both at test time and runtime, this work
delivers the first concrete piece of the representation-level evidence
that EASA learning-assurance guidance demands, and points to
mechanistic interpretability as a practical building block of future
aviation safety cases.
\end{abstract}


\section{Introduction}\label{sec:intro}

Aviation has long set one of the highest safety assurance standards
among safety-critical industries, targeting no more than $10^{-9}$
catastrophic failure conditions per flight hour for transport-category
aircraft \cite{faa_ac_25_1309_1b_2024,easa_cs25}. Empirical fatal
accident rates of the industry's roughly forty million commercial
flights per year come within one to two orders of magnitude of this
demanding target \cite{icao_safety_report_2025}, an impressive track
record built on engineering processes and safety culture developed
over many decades, and an inspiration to safety-critical engineering
in adjacent fields such as automotive, railway, medical devices, and
industrial automation \cite{perezcerrolaza2024aisafety}.

At its core, the success of the safety record rests on a history of
specifying and understanding every step of the analyzed component,
both for hardware and for software, so that failure modes can be
characterized and behavior verified through design reviews, testing,
and formal analysis.

\begin{figure}[t]
  \centering
  \includegraphics[width=\columnwidth]{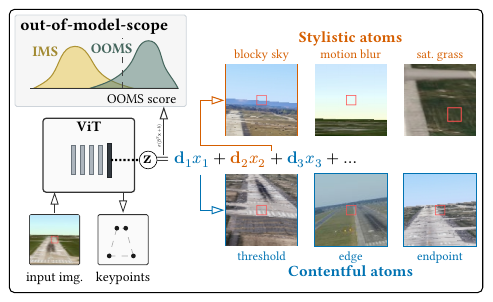}
  \caption{Overview: We disentangle the patch embeddings $\zz$ of a
    vision transformer trained on LARDv2, classify the found atoms
    into \emph{content} and \emph{style}, and use the sparse codes to
    compute an \emph{out-of-model-scope} score.}
  \label{fig:teaser}
\end{figure}

The recent push to introduce data-driven components such as neural
networks to aviation systems breaks this central premise. A neural
network's behavior is learned from data, rather than specified by an
engineer, and traceability and formal analysis can not directly be
applied. Therefore, in practice the assurance task for a neural
network component is often reduced to evaluation on held-out data.
However, this can not be sufficient for an assurance case. DO-178C
\cite{do178c}, which applies to all commercial software-based
aerospace systems, states that test-based evaluation can only be one
part of a solution: verification is not simply testing, and testing
alone cannot show the absence of errors. For data-driven components,
this concern is particularly acute. Performance on a finite test set
may say little about behavior on inputs the test set does not cover;
we cannot readily distinguish a model that has learned a generalizing
mechanism from one that has latched onto spurious correlations purely
from test data \cite{geirhos2020shortcut}.

To address this gap in current software assurance for aviation, EASA
has proposed guidance for Level 1 and 2 machine learning applications
\cite{easa2024ml}. This guidance organizes assurance around a W-shaped
lifecycle with \emph{learning assurance} as a central pillar, and
identifies three runtime monitoring targets: the model \emph{input}
\cite[anticipated EXP-05]{easa2024ml}, the model \emph{output}
\cite[anticipated EXP-06, EXP-07]{easa2024ml}, and the model's
internal \emph{situation representation} \cite[HF-01, HF-03,
anticipated HF-05]{easa2024ml}. Input and output monitoring have each
seen substantial progress in the past. For example, defining the
\emph{operational design domain} (ODD) enables monitoring input
conditions under which the system is qualified to operate. For output,
uncertainty estimation and integrity monitoring provide runtime
checks, and \emph{out-of-distribution} (OOD) detection can be
implemented by comparing outputs of an ensemble of models. For
vision-based landing specifically, prior work has also introduced
integrity checks based on geometric consistency of the predicted
keypoints \cite{previous2025}. However, monitoring the situation
representation itself, i.e., the internal features a neural network
computes to bridge input and output, remains a critical gap which we
aim to address in this work.

Specifically, we introduce the toolkit of \emph{mechanistic
interpretability} \cite{elhage2022superposition,bricken2023sae} to
close the assurance gap between input and output, and apply it to the
keypoint regression task arising in the vision-based landing setting.
This approach aims to directly address EASA's guidance for monitoring
the model's situation representation. Unfortunately, a full
mechanistic account of a deep neural network currently remains out of
reach, since modern models are too high-dimensional, too non-linear,
and too entangled to admit a full white-box explanation of every
decision \cite{sharkey2025openproblems}. Instead, we aim to define and
test \emph{necessary conditions} which provide evidence that a trained
model relies on solid underlying principles to solve the task at hand,
rather than exploiting spurious correlations. More concretely, we
require that the model composes its internal situation representation
as a union of two distinct types of embeddings: task-relevant
geometric structure intrinsic to the keypoint regression task, and
incidental stylistic factors such as color grading, haze, or lighting,
which can help reduce the training loss for a fixed dataset but do not
generalize to novel inputs. Demonstrating that the model correctly
decomposes an input into these two categories of features, and that
the model's predictions rely mostly on the first kind, provides the
foundation of our mechanistic analysis.

Building on this decomposition, we further introduce the new notion of
\emph{out-of-model-scope}, a runtime assurance artifact that
continually verifies that the model has found a sufficient number of
task-relevant features in the input, and allows discarding the
prediction otherwise.

In summary, our contributions are as follows:
\begin{itemize}
  \item We propose a simple and efficient adaptation of the vision
    transformer architecture for runway keypoint regression. We show
    that this architecture is easy to train on the LARDv2 dataset,
    and use the resulting models as a testbed for our mechanistic
    analysis (\Cref{sec:model}).
  \item We decompose the internal feature space of our vision
    transformer into contentful geometric structure and incidental
    stylistic factors, establishing the foundation for mechanistic
    interpretability and runtime assurance
    (\Cref{sec:ksvd,sec:content-style}).
  \item We validate the decomposition through qualitative
    visualization and quantitative analysis of head weights,
    confirming that the regression relies mostly on contentful
    features (\Cref{sec:visualization,sec:head-weight}).
  \item We introduce \emph{out-of-model-scope} monitoring,
    complementing existing input and output monitoring with direct
    monitoring of the model's situation representation, and propose a
    concrete model (\Cref{sec:ooms}).
\end{itemize}

\section{Preliminaries}\label{sec:prelim}

This section reviews the three foundational building blocks on which
the paper relies: (i) the keypoint regression task itself which we aim
to build an assurance case for, (ii) the vision transformer
architecture and regression head that allow for mechanistic analysis,
and (iii) the linear representation theory that theoretically motivates
our mechanistic analysis.

\subsection{The Keypoint Regression Task}
In this work, we focus on the pose estimation sub-problem of a
vision-based landing system. We assume that a computer vision model
receives a single image of the target runway, taken during runway
approach. The goal is then to find, as precisely as possible, the
location of runway keypoints in the image. These keypoints are
typically chosen as the runway corners or other runway markers.
Cross-referencing the estimated two-dimensional keypoint projection
locations with the known geometry of the runway lets us set up an
optimization problem to compute the camera pose, a crucial component
of a vision-based landing system.

Typically, input images are taken from multiple kilometers away so
that the runway appears very small, perhaps less than ten pixels wide.
Therefore, spatial accuracy in the keypoint regression is very
important. For a typical runway approach and camera choice, errors
exceeding two pixels may lead to pose estimate errors of hundreds of
meters, particularly in the alongtrack direction.

We highlight that keypoint regression is fundamentally different to
most other computer vision tasks. Much research in the past decade has
focused on either classification, detection via bounding boxes, or
semantic embedding of input images. In all of these, pixel-level
spatial information is typically discarded as it is not necessary to
solve the task. For example, convolutional neural network
architectures usually discard spatial information using a ``pooling''
operation which aggregates channel-wise information across all spatial
dimensions.

Nonetheless, previous solutions that do use existing spatial structure
for keypoint regression have been proposed in the literature
\cite{xu2022vitpose,sun2018integral}. In particular, in the domain of
human pose estimation, which aims to regress the location of joint
angles, the \emph{softargmax operator (SAM)} has been proposed as a
particularly simple model head for convolutional neural networks
(CNNs) to solve the keypoint regression task. SAM replaces the usual
CNN head, including the pooling layer, thereby preserving existing
spatial information and directly producing a continuous and
differentiable output for image coordinates
\cite{levine2016endtoend,luvizon2019softargmax}. More precisely, SAM
works as follows. Consider a spatial activation map
$\mm{H} \in \reals^{H' \times W'}$ which may arise from selecting a
single channel of a convolutional backbone output. SAM then computes
weighted averages of spatial positions, where the weights are given by
the normalized activation values:
\begin{equation}
  \hat{u} = \sum_{i=1}^{H'} \sum_{j=1}^{W'}
    j \cdot \softmax(\mm{H})_{i,j}
\end{equation}
and similarly for $\hat{v}$. Here $\hat{u}$ denotes the estimated
horizontal image coordinate and $\hat{v}$ the estimated vertical image
coordinate. This computation produces continuous coordinate estimates
with sub-pixel precision while maintaining differentiability, allowing
end-to-end training of the network.

In this work we use and adapt the SAM operator to the vision
transformer architecture to solve the keypoint regression task.

\subsection{Vision Transformers and Patch Embeddings}\label{sec:vit}
The SAM operator has previously been adapted to the keypoint
regression task in the visual landing setting with a CNN as the
backbone \cite{previous2025}. While this setup has been shown to
provide good regression performance, the CNN backbone is not readily
available for mechanistic analysis. The key difficulty in interpreting
CNNs lies in any embedding being an entangled summary of its receptive
field. Although many methods for interpreting CNN models have been
proposed, they typically do not rely on directly analyzing the
embedding space; rather, they rely on interpretations in the input
space \cite{samek2021explaining}.

In contrast to CNNs, a related class of models called vision
transformers (ViTs) \cite{dosovitskiy2021} has been shown to allow for
direct semantic analysis of a model's internal representation
\cite{caron2021emerging,bricken2023sae}, enabled by the \emph{linear
representation theory} \cite{park2024linear}. Since we are interested
in understanding the model's internal representation as a proxy of the
situation representation, in this work we replace the CNN backbone
with that of a ViT. We briefly review the key components of ViTs
required for our analysis and further motivate the analysis of
semantic features in the next section.

ViTs partition an input image into a grid of non-overlapping patches.
A standard ViT formulation such as DINOv2-S \cite{oquab2024}, which we
adapt in this work, takes an input image of size $224 \times 224$ and
partitions it into a $16 \times 16$ grid of $14 \times 14$ pixel
patches. Each patch is embedded into a representation vector which
represents a single token in the transformer architecture
\cite{vaswani2017attention}. The tokens are then processed by the ViT
backbone, producing again $16 \times 16$ output patch embeddings,
which we denote $\zz_p \in \reals^{384}$ for flattened patch index
$p \in \{1, \dots, N\}$. These output patch embeddings provide the
model's patch-level internal representation of the input image and
will serve as the target of our analysis.

\subsection{Linear Representations and Sparse
  Disentanglement}\label{sec:superposition}

Although we have access to the complete internal representations
$\zz_p$ produced by the model, these high-dimensional vectors do not
immediately reveal their semantic content. In other words, we cannot
analyze how the model decomposes the input into interpretable features
such as runway markings, weather conditions, and environment textures.

One might suspect that each dimension in the vector $\zz_p$
corresponds to a distinct, interpretable feature; perhaps a large
value in the first dimension indicates a blue sky, the second
indicates cloud coverage, and so forth. However, it has been shown
that large transformer models do not support such monosemantic
interpretations of axis-aligned features \cite{gurnee2023finding}.

Instead, the linear representation theory \cite{park2024linear}
suggests that for transformer models, which repeatedly recombine
activations with linear operations, semantic concepts correspond to
directions in the embedding space. Rather than activating a single
dimension to encode ``blue sky,'' the model adds an additional
direction to the representation vector. Each patch representation
$\zz_p$ can thus be viewed as a combination of such directional
components, which allows us to assume the form
\begin{equation}
  \zz_p = x_1 \vv{d}_1 + x_2 \vv{d}_2 + \dots
        + x_{n_\dicts} \vv{d}_{n_\dicts}
  \label{eq:Z-eq-D-X}
\end{equation}
for a number of directions $\vv{d}_{(\cdot)}$ and sparse codes
$x_{(\cdot)}$. Notably, the linear representation hypothesis further
assumes a natural sparsity: for a given $\zz_p$, most sparse codes
$x_1, x_2, \dots, x_{n_\dicts}$ will be equal to zero, and only nonzero
for a small number of directions.

To proceed with our analysis, and based on this assumption, our goal
is now to simultaneously determine the set of directions
$\mm{D} = [\vv{d}_1, \dots, \vv{d}_{n_\dicts}]$ that serve as the
basis of all output patch embeddings $\zz_p$, and simultaneously map
each $\zz_p$ onto its sparse code constituents
$x_1, x_2, \dots, x_{n_\dicts}$.

This problem is called \emph{dictionary learning}, and multiple
families of algorithms exist. Sparse autoencoders (SAEs) relax the
sparsity constraint to regularization penalty and learn an
encoder-decoder network via gradient descent, yielding monosemantic
atoms on language and vision transformers
\cite{bricken2023sae,cunningham2023sae}. K-SVD \cite{aharon2006ksvd}
is the classical alternative, which we adopt in this work. The
algorithm alternates between sparse coding via Matching Pursuit
\cite{mallat1993matching} and a closed-form dictionary update. Given a matrix
of activations $\mm{Z} \in \reals^{384 \times n_\patches}$ assembled
from patch embeddings $\zz_p$, K-SVD computes (i) an overcomplete
dictionary $\mm{D} \in \reals^{384 \times n_\dicts}$ of $n_\dicts$
dictionary elements (directions) with $n_\dicts > 384$, and (ii)
sparse codes
$\mm{X} \in \reals^{n_\dicts \times n_\patches}$ satisfying
\begin{equation}
  \min_{\mm{D}, \mm{X}}
    \| \mm{Z} - \mm{D} \mm{X} \|_F^2
  \quad \text{s.t.} \quad
    \| \xx_{(\cdot)} \|_0 \le n_\nnz
  \label{eq:ksvd}
\end{equation}
for all columns $\vv{x}_{(\cdot)}$ of $\mm{X}$ and a target number
$n_\nnz$ of nonzero sparse codes per embedding. In the remainder of
the paper, we refer to these learned dictionary elements $\dd_j$
simply as \emph{atoms}, since they are the basic semantic units of the
sparse decomposition. This disentanglement step provides the
foundation of our analysis of the model internals by breaking the
patch embeddings into their monosemantic components. If we can then
map each atom $\dd_j$ to a human-interpretable representation, or
otherwise categorize the different atoms, this gives us a direct way
to analyze the model's internal situation representation.

\section{Disentangling Content and Style}\label{sec:methods}

With the previous section establishing dictionary learning as a way to
disentangle a model's internal representation into monosemantic
components, we can now pursue our original goal: analyzing the
situation representation inside a computer vision model for
vision-based landing.

Concretely, we train a simple vision transformer model for runway
keypoint regression by combining a ViT backbone together with the SAM
operator. To our knowledge, this application of SAM on a ViT backbone
has not been previously presented in the literature. However, we argue
that the combination provides a simple and efficient approach to
keypoint regression that is easy to train and interpret. We train the
model on the LARDv2 dataset \cite{bougacha2026lardv2}, which comprises
runway approach images generated from two simulators and two satellite
image sources. The trained model then serves as the basis of our
analysis.

We next decompose the model's output patch embeddings into sparse,
semantically meaningful atoms $\dd_j$, the learned dictionary
elements, using K-SVD. This decomposition gives us a window into the
model's situation representation by exposing the internal features
that bridge input and prediction. The rest of our work rests on the
hypothesis that these atoms fall into two categories: \emph{contentful}
atoms that are directly related to solving the task itself, and
\emph{stylistic} atoms that are not inherent to the task but may still
help reduce the loss through spurious correlation. Contentful atoms
may correspond to geometric features of the runway environment itself,
such as runway edges, threshold markings, and horizon lines, which are
intrinsic to the task of keypoint localization. Stylistic atoms
capture incidental visual properties such as color grading,
atmospheric haze, and time-of-day lighting.

This is the bridge to the assurance case: for a model to be assurable,
we argue we must be able to check both that it distinguishes
contentful from stylistic features, and that its output relies mostly
on contentful features.

We therefore categorize each atom found by K-SVD: if the atom's
activation pattern is invariant across the four visual subsets of
LARDv2 we categorize it as contentful, and stylistic otherwise, and
propose this as the foundation of our mechanistic analysis. We
validate this classification in two complementary ways: qualitatively,
by visualizing representative atoms, and quantitatively, by measuring
how much of the regression head's reliance falls on contentful rather
than stylistic atoms.

With the content-style separation validated, we then define a new
runtime assurance notion, \emph{out-of-model-scope} (OOMS),
complementary to both ODD and output-side OOD. The intuition is that
an input may lie well inside the operational design domain and draw no
ensemble disagreement, yet the model may still fail to identify the
task-relevant semantic features; in other words, it is
out-of-model-scope. In this case the prediction should be flagged and
discarded.

The following sections make the proposed mechanistic pipeline precise:
we introduce the construction of a ViT-based keypoint regression model
(\Cref{sec:model}), its K-SVD decomposition (\Cref{sec:ksvd}), a
content-style classifier (\Cref{sec:content-style}), a qualitative
atom visualization (\Cref{sec:visualization}), a quantitative
head-weight readout (\Cref{sec:head-weight}), and a runtime
out-of-model-scope classifier
(\Cref{sec:ooms,sec:ims-lr}).

\subsection{Runway Keypoint Regressor}\label{sec:model}

Our keypoint regressor combines a (potentially pretrained) vision
transformer backbone with the soft-argmax operator from
\Cref{sec:vit}. We use DINOv2-S \cite{oquab2024} as the backbone.
Given a $224 \times 224$ input image, it produces $N = 256$ output
patch tokens $\zz_p \in \reals^{n_\dimst}$ with $n_\dimst = 384$,
along with one CLS token. The crucial property for our purposes is
that each patch token remains tied to a fixed image location. Unlike a
pooled global representation, the output token grid still preserves
the spatial structure required for keypoint regression.

To predict the $n_\kp = 4$ keypoints, we apply SAM not to a
convolutional activation map, but directly to a linear map of the ViT
patch tokens. For each keypoint $k$, a learned map
$\vv{w}_k \in \reals^{n_\dimst}$ assigns a scalar logit
$\vv{w}_k^\top \zz_p$ to every patch token $\zz_p$. After a softmax
over patches, these logits define a spatial weighting of patch center
coordinates, and SAM returns the weighted average of these centers as
\begin{equation}
  (\hat{u}_k, \hat{v}_k) = \sum_{p=1}^N \alpha_{p k} \vv{c}_p,
  \quad
  \alpha_{p k} = \softmax(\vv{w}_k^\top \zz_p).
  \label{eq:softargmax}
\end{equation}
Here $\vv{c}_p \in [0, 1]^2$ is the (normalized) center of patch $p$
on the regular $16 \times 16$ grid, so the output is directly in
image-relative coordinates.

We note that this head introduces only $n_\kp \, n_\dimst \approx
1.5\,\mathrm{k}$ trainable parameters on top of the backbone through
the linear map
$\mm{W} = [\vv{w}_1, \vv{w}_2, \vv{w}_3, \vv{w}_4]^\top$ since the SAM
operator itself is parameter-free. That minimalism is deliberate: the
head is too small to hide substantial task-specific computation, so
the burden of solving the regression problem must sit in the patch
embeddings $\zz_p$ themselves. Those embeddings are therefore the
natural and exclusive target of the mechanistic analysis that follows.

Having constructed the model, we are ready to train it using a dataset
of labeled runway approach images. Specifically, we consider the
LARDv2 dataset which comprises a large number of images taken during
runway approach. However, LARDv2 includes two structural challenges
which we need to resolve before the training.

\subsubsection*{Runway L/R/C disambiguation cue}\label{sec:cue}
LARDv2 samples contain a structural label ambiguity that we must
resolve in order for the keypoint prediction task to be well-posed.
Many samples contain multiple runways in the image; in fact,
parallel-runway airports contribute roughly $64.3\%$ of the dataset,
and at these airports two or three physical strips can project into a
single onboard image. Without additional information, the model would
be asked to solve the impossible task of guessing which runway's
keypoints to locate. In the case of parallel runways, we therefore
expose the runway choice to the model by painting a $7 \times 7$
colored block in the top-left $14 \times 14$ patch at both training
and evaluation time: red for left, blue for right, green for center,
no marking when the runway has no L/R/C suffix. The cue occupies a
fraction of one patch token and leaves the remaining $255$ tokens
untouched; for dictionary fitting and per-patch atom analyses we
exclude the top-left patch so that no atom is trained to encode the
cue itself. A pixel-level cue is a pipeline-simplicity choice that
keeps pilot intent on the same channel as the image in the frozen
DINOv2 interface; a flight-ready system may provide the L/R/C
selection through a dedicated embedding path rather than painting the
image.

\subsubsection*{BOGO crop policy}\label{sec:bogo}
LARDv2 images vary widely in the fraction of the frame occupied by the
runway. Training directly on the raw frames would make training slow
and expensive, and force the network to simultaneously learn runway
localization at arbitrary scales and ignore large regions of sky or
terrain. Instead, we assume the existence of a runway-finder stage
that can pre-crop the full image to a smaller window containing the
runway. We simulate this runway finder model through a BOGO crop
process, named after the infamous bogosort algorithm, and essentially
mirroring rejection sampling. We randomly sample crop proposals of
size $224 \times 224$ pixels until a proposal contains the runway
bounding box with a margin. If the runway does not fit in a $224
\times 224$ pixel crop, we first downsample the image until it does
fit. This process ensures the runway is always present and occupies a
meaningful fraction of the input while maintaining minimal bias (as
opposed to e.g. a center crop).

The BOGO crop construction also has a natural complement:
\emph{inverse-BOGO} crop samples a window from the same image such
that all four runway corners lie outside the crop. This yields an
image for which the keypoint task is unsolvable by construction. We
will reuse this inverse-BOGO construction in \Cref{sec:ims-eval} as
the out-of-model-scope negative class.

\subsection{Sparse Decomposition via K-SVD}\label{sec:ksvd}

Having trained the model, the next goal is to disentangle its output
patch embeddings $\zz_p$. Following \Cref{sec:superposition}, we use
K-SVD and Matching Pursuit to solve \cref{eq:ksvd} under the
per-sample sparsity budget $n_\nnz$ to compute a dictionary
$\mm{D} \in \reals^{n_\dimst \times n_\dicts}$ that serves as a set of
monosemantic basis vectors for the model's situation representation.

At inference, we use this same dictionary in two ways. First, each
patch embedding $\zz_p$ is sparse-coded individually, which gives the
per-patch atom activations used in the content/style analysis. Second,
for each image sample we form an attention-pooled embedding summary
in two steps. The per-keypoint pooled embedding
\begin{equation}
  \vv{s}_k = \sum_p \alpha_{p k} \zz_p
  \label{eq:s-k}
\end{equation}
combines patch embeddings with the same SAM weights $\alpha_{p k}$
that drive the regression head for keypoint $k$, and the image-level
summary
\begin{equation}
  \vv{s} = (1 / n_\kp) \sum_{k=1}^{n_\kp} \vv{s}_k
  \label{eq:s}
\end{equation}
averages across keypoints. Since as in \cref{eq:Z-eq-D-X} we assume
$\zz_p = \mm{D} \vv{x}_p$ with a sparse $\vv{x}_p$, structurally any
weighted sum over patch embeddings will have the form
$\sum w_p \zz_p = \mm{D} \sum w_p \vv{x}_p$. Therefore, by
sparse-coding the summary we hope to directly recover all important
sparse codes for any keypoint prediction at once. In \Cref{sec:ooms}
we will use these sparse codes for the out-of-model-scope classifier.

\subsection{Content vs.~Style via Cross-Subset
  Invariance}\label{sec:content-style}

Having constructed a way to map any input image to its sparse code
activations in the model representations, we now aim to understand the
role of individual units. To this end, the LARDv2 dataset gives us a
natural way to separate content from style. It comprises four subsets
(\emph{xplane}, \emph{ges}, \emph{arcgis}, \emph{bingmaps}) that
depict the same underlying task, runway approach, but with noticeably
different visual rendering. This gives rise to a natural question for
each atom: does the atom activate consistently across all four
subsets, or is its activation predominantly confined to a single
rendering style?

We use this structure as the basis of our distinction of
\emph{contentful} and stylistic atoms. If an atom is active at roughly
the same rate in all four subsets, we treat it as contentful. Such an
atom is responding to structure that is shared across domains, such as
runway geometry, markings, or other approach-relevant layout cues. If,
instead, an atom is concentrated in one or two subsets, we treat it as
\emph{stylistic}, since it is more plausibly tracking
simulator-specific appearance such as texture, color palette, or
rendering artifacts.

Formally, for atom $\dd_j$ and data subset $s$ (\emph{xplane},
\emph{ges}, etc.), let $\mathcal{P}^{s}$ be the set of non-cue patches
from all images in subset $s$. Note that the size $|\mathcal{P}^{s}|$
may differ across subsets if the subsets contain different numbers of
images. We then define the activation rate $p_j^s$ of atom $j$ on
subset $s$ as the fraction of those patches whose sparse code on atom
$j$ is nonzero:
\begin{equation}
  p_j^{s} = \frac{|\{p \in \mathcal{P}^{s} : x_{j p} \ne 0\}|}
                 {|\mathcal{P}^{s}|}.
  \label{eq:rate}
\end{equation}
We can summarize how much this rate varies across subsets with the
coefficient of variation
\begin{equation}
  \mathrm{CV}_j = \frac{\operatorname{std}_s (p_j^{s})}
                       {\operatorname{mean}_s (p_j^{s})},
  \label{eq:cv}
\end{equation}
which is small when the atom fires uniformly across subsets,
indicating contentful atoms, and large when it is subset-specific.

The coefficient of variation lets us now classify each atom as
contentful or stylistic: we split atoms at the median CV, calling
atoms with $p_j^s$ below the median contentful and those with
$p_j^s$ above the median stylistic. This choice is deliberately simple
and introduces no tuned threshold. More importantly, the criterion is
independent of the regression loss. We do not call an atom contentful
because it helps the predictor; we call it contentful because it is
stable across visual domains that preserve the same underlying task.
Later sections then check whether this unsupervised split is
meaningful, first by visualizing representative atoms and then by
measuring how strongly the regression head relies on each side.

\subsection{Atom Visualization Protocol}\label{sec:visualization}

The cross-subset invariance criterion gives us a quantitative split
between contentful and stylistic atoms. Nonetheless, it is still
useful to check whether that split aligns with human judgment. To do
so, we visualize representative atoms by showing the image patches on
which they activate most strongly.

For each atom $\dd_j$, we collect the patches with the largest sparse
coefficients $x_j$ across the validation set and render them at native
resolution together with a small amount of surrounding context. This
lets us inspect what visual pattern an atom is actually responding to,
rather than relying on the CV score alone.

If the classification is meaningful, content atoms should repeatedly
activate on runway structure that a human would also regard as
task-relevant, such as edges, threshold markings, or tarmac. Stylistic
atoms, in contrast, should respond to subset-specific appearance, such
as synthetic rendering artifacts, satellite color palettes, or texture
grain. The full visualization appears in \Cref{sec:atom-viz}.

\subsection{SAM Reliance Score}\label{sec:head-weight}
Visualizing contentful and stylistic atoms gives qualitative insights
into the model's situation representation. Next, we introduce a
quantitative measure of how strongly the regression head relies on
each atom. The goal is to compute an empirical score for each atom
$\dd_j$ that reflects how much it contributes, on average, to the
keypoint regression output.

To construct this score we recall that the regression output
$(\hat{u}_k, \hat{v}_k)$ for a single keypoint $k$ can be computed as
\begin{equation}
  (\hat{u}_k, \hat{v}_k)
    = \SAM\!\left( \vv{w}_k^\top
        [\vv{z}_1, \dots, \vv{z}_{n_p}] \right).
  \label{eq:SAM-1}
\end{equation}
Further, rewriting each $\zz$ as
\begin{equation}
  z = [\vv{d}_1, \dots, \vv{d}_{n_\dicts}]
      \begin{bmatrix} x_1 \\ \vdots \\ x_{n_\dicts} \end{bmatrix}
    = \mm{D} \vv{x}
\end{equation}
and inserting into \cref{eq:SAM-1} reveals
\begin{equation}
  (\hat{u}_k, \hat{v}_k)
    = \SAM\!\left( \vv{w}_k^\top \mm{D}
        [\vv{x}_1, \dots, \vv{x}_{n_p}] \right).
\end{equation}
Recalling that the sparse codes are zero most of the time, we can
characterize the average contribution of an atom $\dd_j$ to the
regression output using three factors:
\begin{enumerate}
  \item How aligned $\dd_j$ is with the head's linear projection
    $\vv{w}_k$ for each keypoint $k$, i.e., $\vv{w}_k^\top \dd_j$.
    Considering all keypoints at once we compute
    $\| \mm{W} \dd_j \|_2$ with
    $\mm{W} = [\ww_1, \dots, \ww_{n_\kp}]^\top$.
  \item How often an atom is active, i.e., how frequently the sparse
    code $x_j$ for atom $\dd_j$ is nonzero when averaged over all
    patches of all images. We write $\Prob_{\zz}[x_j \ne 0]$.
  \item How strongly an atom activates provided that it is active at
    all. We write $\Expect_z[|x_j| \mid x_j \ne 0]$.
\end{enumerate}
We then compute the reliance score for each atom $\dd_j$ as
\begin{align}
  \mathrm{score}_j
    &= \| \mm{W} \dd_j \|
       \times \Prob_z[x_j \ne 0]
       \times \Expect_z[|x_j| \mid x_j \ne 0] \notag \\
    &= \| \mm{W} \dd_j \| \times \Expect_z[|x_j|]
  \label{eq:effective}
\end{align}
where the expectation and probability are empirically evaluated over
the patches of all image samples.

We conjecture that an assurable model should place most of its
regression attention on contentful atoms rather than stylistic ones.
We therefore aggregate the scores across all atoms to compute the
fraction of total reliance that falls on the contentful side:
\begin{equation}
  \frac{\sum_j \mathrm{score}_j \times \mathds{1}[\dd_j\text{ is contentful}]}
       {\sum_j \mathrm{score}_j}
\end{equation}
and report the results in \Cref{sec:content-style-results}.

This quantitative analysis concludes the inspection of the constructed
model's internal activations. Next, we will introduce the additional
notion of \emph{out-of-model-scope detection} as a novel approach to
runtime assurance, building on top of the content/style split
introduced in this section.

\section{Out-of-Model-Scope Detection}\label{sec:ooms}

EASA's runtime monitoring guidance distinguishes three targets: the
input, the output, and the \emph{situation representation}. The first
two already have established counterparts. Input-side monitoring
yields the \emph{operational design domain} (ODD), and output-side
monitoring yields uncertainty estimation, integrity checks, and
ensemble-based \emph{out-of-distribution} (OOD) detection. What
remains missing is a monitor for the situation representation itself.

We call this notion \emph{out-of-model-scope} (OOMS). An input is
\emph{in-model-scope} (IMS) if the model's internal representation
contains the contentful features on which the regression head has been
shown to rely. If those features are absent, the sample is OOMS and
the prediction should be discarded.

This criterion is different from both ODD and output-side OOD. ODD
asks whether the input is a plausible member of the operational
environment, and output-side OOD asks whether the model's outputs look
statistically atypical or mutually inconsistent. OOMS instead asks
whether the internal situation representation contains the
task-relevant semantic structure needed to support the prediction.

This distinction matters because a sample may look perfectly valid at
the pixel level and still produce a numerically confident output,
while failing to activate the contentful atoms that the head actually
uses. OOMS is meant to catch exactly this failure mode.

\subsection{OOMS Classifier via Logistic Regression}\label{sec:ims-lr}

Having defined OOMS qualitatively in \Cref{sec:ooms}, we now turn the
criterion into a runtime classifier whose job is to read off, from the
atoms active on a single image, whether enough of the relied-on
structure is present to trust the prediction.

We obtain those activations by sparse-coding the image-level summary
$\vv{s}$ from \cref{eq:s-k,eq:s} against the K-SVD dictionary $\mm{D}$,
using Matching Pursuit at the same sparsity $n_\nnz = 8$ as elsewhere.
This yields a coefficient vector
$\vv{x}(\vv{s}) \in \reals^{n_\dicts}$ supported on the same atom
vocabulary as the content/style analysis, but evaluated at the image
level on which the head actually operates. We then take the binary
support
\begin{equation}
  a_j = \mathds{1}[x_j(\vv{s}) \ne 0] \in \{0, 1\},
  \quad j = 1, \dots, n_\dicts
\end{equation}
as the input feature. Using the continuous magnitudes
$|x_j(\vv{s})|$ in place of $a_j$ also yields a working classifier,
but the binary support keeps the resulting model substantially simpler
(both to fit and to read off), and we adopt it for the rest of the
paper.

Given $\vv{a}$, the classifier outputs the IMS probability
\begin{equation}
  \hat{p}_\mathrm{IMS}(\vv{a})
    = \sigma\!\left( b
       + \sum_{j=1}^{n_\dicts} \beta_j a_j \right),
  \label{eq:lr}
\end{equation}
with $n_\dicts = 512$ the full dictionary size. We fit $b$ and
$\bbeta$ by minimizing binary cross-entropy with an $L_1$ penalty
under sign constraints,
\begin{equation}
  \min_{b \le 0,\ \bbeta \ge 0}
    \BCE(\bbeta, b) + \lambda \| \bbeta \|_1.
  \label{eq:lr_l1}
\end{equation}
The constraints directly encode intended semantics of the detector.
The bias constraint $b \le 0$ forces the default prediction to be
OOMS, and the positivity constraint $\beta_j \ge 0$ ensures that each
active atom can only push the probability toward IMS, never against
it. Under positivity, $\| \bbeta \|_1 = \sum_j \beta_j$, so
\cref{eq:lr_l1} remains a smooth box-constrained convex problem that
we solve with L-BFGS-B \cite{byrd1995lbfgsb}. The $L_1$ term drives
most coefficients to zero, so the fit automatically selects an active
subset of $n_\dictssel$ atoms whose number is set by $\lambda$. We can
then inspect if the selected atoms correspond to contentful or
stylistic atoms.

\begin{figure*}[t]
  \centering
  \includegraphics[width=\textwidth]{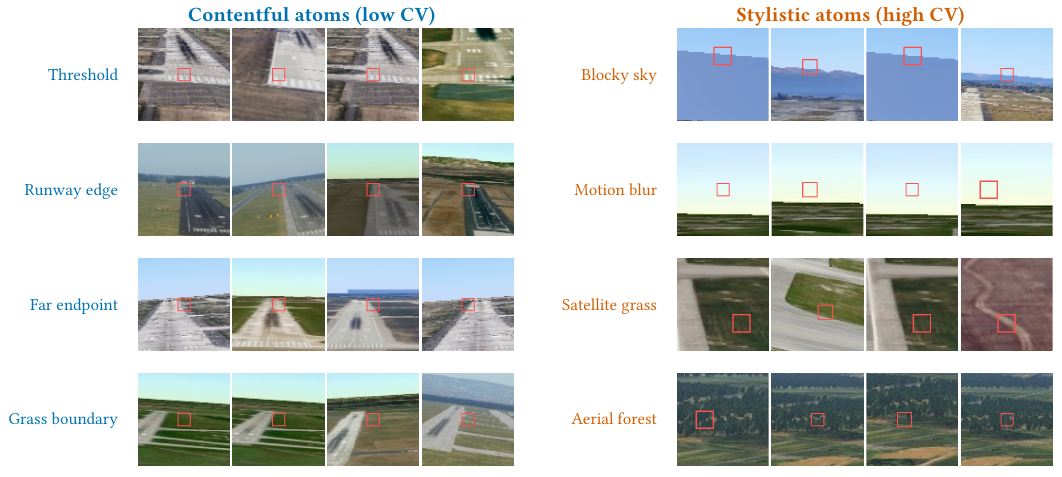}
  \caption{Top-activating patches for four contentful atoms (left)
    and four stylistic atoms (right); three-patch context, target
    outlined in red.}
  \label{fig:atoms}
\end{figure*}

\section{Experiments}\label{sec:experiments}

\subsection{Model Training and Regression
  Performance}\label{sec:reg-perf}

\subsubsection*{Training Setup}\label{sec:reg-setup}
All experiments use the LARDv2 dataset \cite{bougacha2026lardv2} and
its four visual domains, \emph{xplane} (synthetic flight simulator),
\emph{ges} (Google Earth Studio), \emph{arcgis} (ArcGIS satellite),
and \emph{bingmaps} (Bing Maps satellite); all four depict runways
from approach perspectives but with different appearance. This
multi-domain structure underpins the content/style analysis that
follows. Samples for training, validation, and test are generated with
both the BOGO crop of \Cref{sec:bogo}, which guarantees a
runway-containing window, and the top-left L/R/C runway cue of
\Cref{sec:cue}. Keypoint coordinate labels are normalized to $[0, 1]$
relative to the crop before being passed to the loss. However, for
convenience we report un-normalized errors in units of pixels. For
the \emph{pretrained} variant we use a DINOv2-S backbone initialized
from self-supervised pretraining on LVD-142M \cite{oquab2024} and
fine-tune end-to-end for $30$ epochs with AdamW (weight decay
$10^{-4}$), linear warmup for $5$ epochs then cosine decay, gradient
clipping at norm $1.0$, and learning rate $5 \times 10^{-4}$ with the
backbone learning rate reduced by $0.1\times$ so the pretrained
features are perturbed only gently. The head-only component
\cref{eq:softargmax} and the backbone share the same optimizer. The
from-scratch baseline uses the same architecture, optimizer family,
and loss, but is trained from random initialization for twice as many
epochs. The loss is a Huber loss on the $[0, 1]$-normalized keypoint
predictions with breakpoint $\delta = 8 / 224 \approx 0.036$. This
ensures that small residuals are penalized quadratically and large
residuals linearly, reducing the influence of outliers, which we find
necessary for the LARDv2 dataset which contains a number of occluded
or otherwise challenging samples. For each visible keypoint we compute
the Euclidean pixel error between prediction and target, and notably
report the value directly, rather than the squared error, to aid
readability of the reported numbers. We report its mean over all
visible keypoints on the test set, and the median over samples of the
per-sample average visible-keypoint error.

\begin{table}[t]
  \centering
  \caption{Regression performance on LARDv2 (5 seeds).}
  \label{tab:training}
  \begin{tabular}{@{}lcccc@{}}
    \toprule
    & \multicolumn{2}{c}{\textbf{Train}}
    & \multicolumn{2}{c}{\textbf{Test}} \\
    \cmidrule(lr){2-3}\cmidrule(lr){4-5}
    \textbf{Variant}
      & \textbf{MAE [px]} & \textbf{Median [px]}
      & \textbf{MAE [px]} & \textbf{Median [px]} \\
    \midrule
    Pretrained
      & $\mathbf{1.6 \pm 0.2}$ & $\mathbf{1.3 \pm 0.1}$
      & $\mathbf{4.4 \pm 0.1}$ & $\mathbf{1.9 \pm 0.0}$ \\
    Scratch
      & $1.7 \pm 0.1$ & $1.5 \pm 0.1$
      & $7.4 \pm 0.2$ & $3.2 \pm 0.1$ \\
    \bottomrule
  \end{tabular}
\end{table}

\subsubsection*{Results}\label{sec:reg-results}
\Cref{tab:training} summarizes the keypoint regression performance of
the pretrained and from-scratch variants. The pretrained model reaches
a test-set median pixel error of $1.91$\,px ($\pm 0.03$, $n=5$), which
is typically acceptable for the downstream PnP problem. The mean
pixel error is nearly two and a half times higher at $4.39$\,px
($\pm 0.14$), symptomatic of a long tail of heavily occluded or
partially out-of-frame samples that the median intentionally
discounts. The from-scratch baseline trails on both metrics (median
$3.19$\,px ($\pm 0.07$), MAE $7.45$\,px ($\pm 0.17$), $n=5$) despite
training for twice as many epochs; the gap widens at the tail,
exactly where an unpretrained backbone would be expected to struggle.

\subsection{K-SVD Dictionary Fitting}\label{sec:ksvd-fit}

\subsubsection*{Setup}\label{sec:ksvd-setup}
For the K-SVD dictionary of \Cref{sec:ksvd} we collect the
$384$-dimensional output patch embeddings from the trained model on
the test set: one CLS token and the top $4$ patch tokens per image
(ranked by sum of head-attention magnitudes across the four
keypoints), giving approximately $237{,}000$ vectors from the
$47{,}472$ test images. For this and all other analysis, the top-left
(cue) patch is always excluded.

This importance-sampled pool avoids dictionary dominance by the
background-patch mode, which outnumbers the task-relevant patches by
almost two orders of magnitude. We fit $n_\dicts = 512$ atoms at
per-sample sparsity $n_\nnz = 8$ using a batched K-SVD implementation
\cite{aharon2006ksvd,valentin2025dbksvd} with batch size $8192$, $3$
epochs over the pool, and Parallel Matching Pursuit as the inner
solver. The fitted dictionary is held fixed for all downstream
analyses. At inference each token is encoded via Matching Pursuit at
the same sparsity level.

\subsubsection*{Results}\label{sec:ksvd-results}
The fitted $384 \times 512$ dictionary achieves a reconstruction
variance explained above $90\%$ on a held-out batch of the
importance-sampled pool, confirming that $8$ atoms per patch provide a
faithful approximation of the embedding space. All $12{,}152{,}832$
test-set patch tokens are then sparse-coded via Matching Pursuit
against this fixed dictionary, yielding the activation patterns from
which every subsequent analysis is derived.

\subsection{Content vs.~Style}\label{sec:content-style-results}

\subsubsection*{Setup}\label{sec:cs-setup}
We classify each of the $512$ atoms as contentful or stylistic by
thresholding the cross-subset CV \cref{eq:cv} at its median across all
atoms active on the test set. For the head-weight readout
(\Cref{sec:head-weight}) we compute the effective score of
\cref{eq:effective} for every atom and report the fraction of total
effective score lying on contentful versus stylistic atoms.

\begin{table}[t]
  \centering
  \caption{Head-weight split across the dictionary (5 seeds).}
  \label{tab:content-style}
  \begin{tabular}{@{}lcc@{}}
    \toprule
    \textbf{Variant} & \textbf{Contentful} & \textbf{Stylistic} \\
    \midrule
    \textbf{Pretrained}
      & $\mathbf{0.82 \pm 0.05}$ & $\mathbf{0.18 \pm 0.05}$ \\
    Scratch
      & $0.80 \pm 0.04$ & $0.20 \pm 0.04$ \\
    \bottomrule
  \end{tabular}
\end{table}

\subsubsection*{Results}\label{sec:cs-results}
On the pretrained model the head places $81.7\%$ ($\pm 5.4\%$) of its
total effective regression score on contentful atoms and the remaining
$18.3\%$ on stylistic ones (\Cref{tab:content-style}, $n=5$). The
from-scratch baseline lands at $79.8\%$ ($\pm 4.4\%$) content and
$20.2\%$ style. We posit that the good test set performance of each
model variant can be explained by the focus on mostly contentful
atoms. However, we note that for the contentful reliance score both
variants are statistically indistinguishable. In these experiments the
contentful reliance score therefore cannot explain the test-set
performance gap between both variants.

\begin{figure}[t]
  \centering
  \includegraphics[width=\columnwidth]{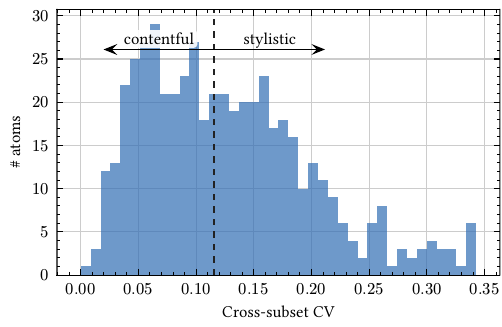}
  \caption{Cross-subset CV distribution across the $512$ atoms;
    dashed line is the median threshold used to split content from
    style.}
  \label{fig:cv-hist}
\end{figure}

\subsection{Atom Visualization}\label{sec:atom-viz}

\subsubsection*{Setup}\label{sec:viz-setup}
We realize the visualization protocol of \Cref{sec:visualization} on
the pretrained then fine-tuned model: for a slightly curated set of
contentful and stylistic atoms we render the top $n_\viz = 4$
activating patches with a three-patch context window, with the target
patch outlined.

\subsubsection*{Results}\label{sec:viz-results}
\Cref{fig:atoms} shows the result. Contentful atoms consistently fire
on patches that depict human-recognizable runway structure: threshold
markings, tarmac edges, and horizon lines. Stylistic atoms, by
contrast, respond to domain-identifying appearance features such as
rendering artifacts or environment textures. This qualitative
inspection supports the CV-based classifier from
\Cref{sec:content-style} without any reference to the regression
loss, and provides the kind of visual evidence that a certification
reviewer can directly inspect.

\subsection{OOMS Detection}\label{sec:ims-eval}

\subsubsection*{Setup}\label{sec:bogo-setup}
For OOMS detection we sample IMS and OOMS images according to
\Cref{sec:ims-lr}: the positive class is the standard BOGO-cropped
LARDv2 test set (runway in frame); the negative class is the
inverse-BOGO construction from \Cref{sec:model}, in which all four
runway corners lie outside the crop so that the keypoint task is
unsolvable by construction. Features are the binary atom-activation
vectors on the non-cue attention-pooled summary from the pretrained
then fine-tuned model, with the summary as defined in
\Cref{sec:ims-lr}. We fit the constrained LR of \cref{eq:lr_l1} on a
70/30 train/evaluation split and sweep the $L_1$ strength $\lambda$ to
trace the sparsity--accuracy frontier.

\begin{figure}[t]
  \centering
  \subfloat[AUROC vs sparsity.\label{fig:ims-auroc}]{%
    \includegraphics[width=0.49\columnwidth]{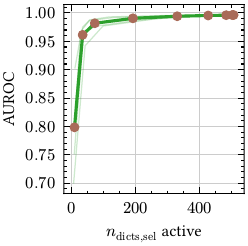}}
  \hfill
  \subfloat[Content fraction of $L_1$-selected atoms.\label{fig:ims-content-frac}]{%
    \includegraphics[width=0.49\columnwidth]{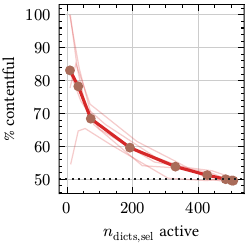}}
  \caption{OOMS detector $L_1$ sweep, $5$ seeds (faint) with mean
    (bold).}
  \label{fig:ims-sweep}
\end{figure}

\subsubsection*{Results}\label{sec:bogo-res}
\Cref{fig:ims-sweep} presents the results of the OOMS analysis.
Picking one illustrative sample from \Cref{fig:ims-auroc} we find that
for one choice of $\lambda = 300$ the constrained logistic regression
model selects $n_\dictssel = 34$ atoms ($\pm 6$, $n=5$) from the
512-atom dictionary, all with positive weights and strictly negative
bias by \cref{eq:lr_l1}. On the held-out 30\% of BOGO and inverse-BOGO
samples the classifier reaches AUROC $0.96$ ($\pm 0.01$) with those
atoms alone. In other words, checking the existence of $34$ out of
$512$ atoms is sufficient to detect with great accuracy if the input
contains a runway or not.

We further analyze the selected atoms for whether they are contentful
or not. We find that at $\lambda = 300$, $78\%$ ($\pm 8\%$) of the
selected atoms are classified as \emph{contentful} by the CV criterion
of \Cref{sec:content-style}, well above the $50\%$ baseline a random
pick would give, since the CV threshold is the median. As we decrease
the $L_1$ regularization parameter, we eventually approach the natural
split of approximately $50/50$ contentful and stylistic atoms, as for
low $\lambda$ all sparse codes are selected.

\section{Discussion and Conclusion}\label{sec:discussion}

We proposed mechanistic interpretability as a technical means to
monitor the \emph{situation representation} of a vision-based aircraft
landing system, filling a gap in current EASA learning-assurance
guidance. The model must learn a situation representation that
decomposes into contentful and stylistic atoms, and the prediction
head must place most of its weight on the contentful atoms. This
yields an assurance artifact a certification reviewer can inspect
directly: the model's output is trustworthy only when the internal
situation representation contains the contentful atoms it demonstrably
relies on. Building on this idea we further propose a runtime monitor
that checks whether an input is \emph{out-of-model-scope}, contrasting
with the input- and output-side monitors EASA already identifies: ODD
and OOD check whether the input is admissible or the output is
statistically typical, whereas OOMS checks whether the internal
representation contains the semantic structure required for the head
to make a prediction. On the BOGO vs.\ inverse-BOGO benchmark the
classifier of \Cref{sec:ims-lr} selects $34$ ($\pm 6$, $n=5$) atoms
automatically, reaches AUROC $0.96$ ($\pm 0.01$), and $78\%$ ($\pm 8\%$)
of its selected atoms carry the content label.

Our OOMS monitor complements rather than replaces conventional OOD
methods such as Mahalanobis-distance monitors or ensemble disagreement.
Distance and ensemble detectors flag that a sample is statistically
unusual; OOMS additionally reports \emph{which} of the model's
interpretable content atoms failed to fire, carrying a direct link to
the EASA situation-representation requirement those detectors do not
supply. Beyond this experiment, the approach gives a
\emph{model-internal} diagnostic for shortcut learning, which the
literature \cite{geirhos2020shortcut} has otherwise addressed mostly
through probe datasets or adversarial evaluations.

Several limitations remain. The content/style classification assumes
the four LARDv2 domains adequately span the space of stylistic
variation; if all four share an artifact (for instance, clear skies),
atoms responding to it would be misclassified as contentful. The
median-CV threshold is a simple choice, and a broader causal account
via per-atom ablation or activation scrubbing is for future work. The
scope is currently a single task and architecture. Steps toward
certification-ready evidence include a formal specification of which
atoms \emph{should} be contentful for a given task, a dictionary
coverage analysis, and stability analysis across retraining seeds.
Future work will integrate this representation-level monitoring with
output-level protection levels \cite{previous2025} for a full runtime
assurance stack.

\bibliographystyle{IEEEtranN}
\bibliography{refs}

\end{document}